\pdfoutput=1

\documentclass[11pt]{article}

\usepackage[final]{acl}

\usepackage{times}
\usepackage{latexsym}

\usepackage{todonotes}
\usepackage{svg}

\usepackage[T1]{fontenc}

\usepackage[utf8]{inputenc}

\usepackage[nopatch=footnote]{microtype}

\usepackage{inconsolata}

\usepackage{anyfontsize}
\usepackage{graphicx}
\usepackage{booktabs}
\usepackage{hyperref}
\usepackage{url}
\usepackage{multirow}
\usepackage{subfigure}
\usepackage{makecell} 
\usepackage{booktabs}
\usepackage{booktabs} 
\usepackage{multirow} 
\usepackage{amsmath} 
\usepackage{graphicx} 

\usepackage{placeins}
\usepackage{enumerate}
\usepackage{enumitem}
\usepackage{lipsum}
\usepackage{xcolor}
\usepackage{booktabs}
\usepackage{multirow}
\usepackage{tabularx}

\definecolor{cvprblue}{rgb}{0.21,0.49,0.74}
\definecolor{citecolor}{HTML}{0071BC}
\definecolor{linkcolor}{HTML}{ED1C24}
\usepackage[capitalize]{cleveref}
\crefname{section}{Sec.}{Secs.}
\crefname{table}{Table}{Tables}
\crefname{figure}{Fig.}{Figs.}
\usepackage[most]{tcolorbox}
\usepackage{float}
\usepackage{xspace}
\tcbset{
  aibox/.style={
    width=474.18663pt,
    top=10pt,
    colback=white,
    colframe=black,
    colbacktitle=black,
    enhanced,
    center,
    attach boxed title to top left={yshift=-0.1in,xshift=0.15in},
    boxed title style={boxrule=0pt,colframe=white,},
  }
}
\newtcolorbox{AIbox}[2][]{aibox,title=#2,#1}

\usepackage{colortbl}

\usepackage{multicol} 
\usepackage[most]{tcolorbox}
\usepackage{float}
\usepackage{xspace}
\usepackage{fontsize}
\usepackage{setspace}
\usepackage{geometry}

\usepackage{listings}
\definecolor{codegreen}{rgb}{0,0.6,0}
\definecolor{codegray}{rgb}{0.5,0.5,0.5}
\definecolor{codepurple}{rgb}{0.58,0,0.82}
\definecolor{backcolour}{rgb}{0.95,0.95,0.92}
\definecolor{mypurple}{RGB}{200,192,248}
\definecolor{mypurpledeep}{RGB}{142,126,240}
\definecolor{mygreen}{RGB}{117,170,156}
\definecolor{myyellow}{RGB}{255,192,0}
\definecolor{myblue}{RGB}{57,143,255}
\definecolor{mygrey}{RGB}{231,230,230}
\definecolor{codey}{RGB}{220,220,170}
\definecolor{coder}{RGB}{206,145,120}
\definecolor{codeb}{RGB}{156,220,254}
\definecolor{codenum}{RGB}{204,204,204}

\lstdefinestyle{mystyle}{
    backgroundcolor=\color{backcolour},   
    commentstyle=\color{codegreen},
    keywordstyle=\color{magenta},
    numberstyle=\tiny\color{codegray},
    stringstyle=\color{codepurple},
    breakatwhitespace=false,         
    breaklines=true,                 
    captionpos=b,                    
    keepspaces=true,                 
    numbers=left,                    
    numbersep=5pt,                  
    showspaces=false,                
    showstringspaces=false,
    showtabs=false,                  
    tabsize=2
}

\usepackage{algorithm} 

\author{
   Justin Xu\thanks{Equal contribution}{$^1$}
   \\\And
   Zhihong Chen$^{*1}$
   \\\And
   Andrew Johnston$^1$
   \\\And
   Louis Blankemeier$^1$
   \\\AND
   Maya Varma$^1$ 
   \\\And
   Jason Hom$^1$ 
   \\\And
   William J. Collins$^1$
   \\\AND
   Ankit Modi$^2$
   \\\And
   Robert Lloyd$^2$
   \\\And
   Benjamin Hopkins$^3$
   \\\AND
   Curtis P. Langlotz$^1$ 
   \\\And
   Jean-Benoit Delbrouck$^1$ 
   \\\AND
   $^{1}$Stanford University \hspace{0.2cm}   $^{2}$University of Arizona   \\$^{3}$\textbf{University of Southern California}\\
   \texttt{\{xujustin,zhihongc,drewj32,langlotz,jbdel\}@stanford.edu}
}

\begin{document}
\title{Overview of the First Shared Task on Clinical Text Generation:\\RRG24 and ``Discharge Me!''}

\maketitle

\begin{abstract}
Recent developments in natural language generation have tremendous implications for healthcare. For instance, state-of-the-art systems could automate the generation of sections in clinical reports to alleviate physician workload and streamline hospital documentation. To explore these applications, we present a shared task consisting of two subtasks: (1) Radiology Report Generation (RRG24) and (2) Discharge Summary Generation (``Discharge Me!''). RRG24 involves generating the `Findings' and `Impression' sections of radiology reports given chest X-rays. ``Discharge Me!'' involves generating the `Brief Hospital Course' and `Discharge Instructions' sections of discharge summaries for patients admitted through the emergency department. ``Discharge Me!'' submissions were subsequently reviewed by a team of clinicians. Both tasks emphasize the goal of reducing clinician burnout and  repetitive workloads by generating documentation. We received 201 submissions from across 8 teams for RRG24, and 211 submissions from across 16 teams for ``Discharge Me!''.
\end{abstract}

\section{Introduction} \label{sec:introduction}
An important application of natural language generation (NLG) in medical artificial intelligence (AI) is radiology report generation (RRG). Specifically, an RRG system can be designed to accept radiology images (\textit{e.g.,} chest X-rays) of a patient and generate a textual report describing the clinical observations in the images. This is a clinically important task, and offers the potential to reduce the repetitive work of radiologists and generally improve clinical communication~\cite{pang_survey_2023}. Existing studies have been conducted using a single dataset, which limits the scale and diversity of the data and results. Therefore, we introduce our first subtask, RRG24, where we curate Interpret-CXR, a large-scale collection of RRG datasets from a variety of different sources (\textit{i.e.,} MIMIC-CXR~\cite{johnson2019mimic}, CheXpert~\cite{irvin2019chexpert}, PadChest~\cite{bustos2020padchest}, BIMCV-COVID19~\cite{vaya2020bimcv}, and OpenI~\cite{openi}). In RRG24, participants generate the Findings and Impression sections from chest X-rays. We then evaluate the generations on common leaderboards with standard and recently proposed metrics. Ultimately, this task aims to benchmark recent progress using common data splits and evaluation implementations.

NLG can also impact discharge documentation by playing a role in generating discharge summaries. Hence, we introduce our second subtask, ``Discharge Me!'', with the primary objective of encouraging NLG systems that alleviate clinician burden when writing detailed discharge summaries. Clinicians play a crucial role in documenting patient progress after a hospital stay, but the creation of concise yet comprehensive Brief Hospital Course (BHC) sections and Discharge Instructions often demands a significant investment of time~\cite{do_augmented_2020, alissa_saving_2021}. These two sections in particular cannot be readily copied from prior notes, and thus must be written from scratch by clinicians who synthesize information from across the patient record~\cite{weetman_what_2021}. This process contributes to clinician burnout and poses operational inefficiencies within hospital workflows~\cite{haycock_improving_2014}. We hypothesize that computer-generated clinical documentation has the potential to more accurately and completely capture a patient’s hospital course while reducing the administrative burden on clinicians, which, in turn, mitigates burnout, streamlines hospital operations, and ultimately improves the quality of care. Thus, in ``Discharge Me!'', participants submit generations of both target sections (BHC \& Discharge Instructions). We evaluate submissions on a common leaderboard and conduct a subsequent manual clinician review to measure clinical alignment of the outputs.

\section{Related Work} \label{sec:related}
\subsection{Radiology Report Generation}
\label{sec:task_a_related}
Recent advances in computer vision (CV) and NLG have shown great potential for the automatic generation of radiology reports. This progress can be summarized from three perspectives:
\begin{itemize}[labelindent=0cm,topsep=0pt,itemsep=0pt,parsep=0pt]
    \item (1) Data: Most relevant studies focus on chest X-rays, mainly owing to the current number of publicly available image-report datasets for this modality (\textit{e.g.,} MIMIC-CXR, PadChest, and OpenI, etc.). Recently, there have also been studies expanding the scope of radiology report generation to other modalities (\textit{e.g.,} computed tomography (CT)~\cite{loveymi_automatic_2021,hamamci_ct2rep_2024} and ultrasound~\cite{zeng_deep_2020,yang2021automatic,huh_breast_2023}).
    \item (2) Methodology: The methods for radiology report generation have evolved from task-specific modeling to pre-training-based approaches. For the former, researchers have incorporated the task priors into the designs of the model architectures~\cite{shin2016learning,zhang2017mdnet,jing2018automatic,chen2020generating,zhang2020radiology,liu2021exploring,delbrouck2022improving,hou2023organ}, whereas for the latter, researchers have performed domain-specific representation learning using vision encoders or have adopted large pre-trained language decoders~\cite{thawkar2023xraygpt,hyland2023maira,tu2024towards}.
    \item (3) Evaluation: One of the largest factors hampering radiology report generation progress is the selection of evaluation metrics. Due to its domain-specific characteristics, simple \textit{n}-gram matching metrics (\textit{e.g.,} BLEU~\cite{papineni2002bleu}, ROUGE~\cite{lin2004rouge}, METEOR~\cite{banerjee_meteor_2005}, and CIDEr~\cite{vedantam_cider_2015}) are sub-optimal choices for this task. However, researchers have proposed various model-based metrics for evaluating the quality of generated reports, such as BERTScore~\cite{zhang2019bertscore}, F1-CheXbert~\cite{smit2020combining}, F1-RadGraph~\cite{delbrouck2022improving}, and GREEN~\cite{ostmeier_green_2024}.
\end{itemize}

\subsection{Discharge Summary Generation} \label{sec:task_b_related}
Previous research has also examined AI technologies for the generation of discharge summaries to alleviate clerical burden for clinicians. For instance, several studies investigated GPT-3.5's and GPT-4's capability to generate discharge notes in tandem with various prompting strategies. In a UK pilot feasibility study, it was observed that a set of 25 AI-generated summaries were all deemed acceptable by general practitioners, compared to 23/25 (92\%) of summaries written by junior doctors~\cite{clough_transforming_2024}. Other studies similarly concluded that these proprietary models exhibit great potential and are able to generate acceptable discharge summaries with minimal misinformation~\cite{kim_patient-friendly_2024, waisberg_gpt-4_2023}. However, despite being able to increase efficiency and reduce the time required for documentation as compared to writing or dictating notes, instances of hallucination or omission of clinically significant facts were observed for certain discharge summaries involving complex surgeries. As such, the factual correctness of these large language models (LLMs) for specific generation tasks could be improved~\cite{williams_evaluating_2024, dubinski_leveraging_2024}.

Based on this, some studies have focused on generating a particular section common to most discharge summaries -- BHC -- optimizing for correctness and faithfulness. The BHC is a succinct summary of a patient's entire journey through the hospital and are embedded within complex discharge summaries. Efforts in compiling large-scale datasets for the generation of these BHC sections~\cite{adams_whats_2021}, including those with synthetic data~\cite{adams_learning_2022}, have led to subsequent contrastive learning methods for aligning generation models~\cite{adams_what_2023}. Finally, methods leveraging heuristics to increase factuality (\textit{e.g.,} retrieval and ontology referencing) have also been developed~\cite{adams_speer_2024, hartman_method_2023}.

Some research has similarly centered on the Discharge Instructions section, sometimes known as the Patient Instructions section. This section is patient-facing and details instructions for the patient to continue their care at home, such as information on diet, therapies, and medications, as well as any details for follow-up appointments. Patient readability of this section is critical, and LLMs could be used to reformulate them into a more patient-friendly language~\cite{zaretsky_generative_2024}. Similar to the BHC, previous work also explored frameworks for the generation of faithful Patient Instructions~\cite{liu_retrieve_2022}.

\section{RRG24: Radiology Report Generation} \label{sec:task_a}
RRG24 was hosted on ViLMedic~\cite{delbrouck2022vilmedic}, a modular framework for vision-language multimodal research in medicine. The library contains reference implementations for state-of-the-art vision-language architectures for medicine and also hosts shared challenges in AI. A total of 201 submissions were received from across 8 teams.

\subsection{Data} \label{sec:task_a_data}
We curated Interpret-CXR, a large-scale collection of RRG datasets from the following five sources: MIMIC-CXR~\cite{johnson2019mimic}, CheXpert~\cite{irvin2019chexpert}, PadChest~\cite{bustos2020padchest}, BIMCV-COVID19~\cite{vaya2020bimcv}, and OpenI~\cite{openi}. The breakdown of Interpret-CXR, including details of the four splits used in RRG24 (Training, Validation, Public Test, and Hidden Test) are reported in Table~\ref{table:task_a_breakdown}.

\begin{table*}[t]
\caption{Dataset Breakdown of Interpret-CXR for RRG24}
\label{table:task_a_breakdown}
\centering
\resizebox{0.90\linewidth}{!}{
\begin{tabular}{@{}lrrrrrrrr@{}}
\toprule
\multirow{2.5}{*}{Dataset} & \multicolumn{2}{c}{Training} & \multicolumn{2}{c}{Validation} & \multicolumn{2}{c}{Public Test} & \multicolumn{2}{c}{Hidden Test} \\ \cmidrule(l){2-9} 
                         & Findings     & Impression    & Findings      & Impression     & Findings      & Impression      & Findings      & Impression      \\ \midrule
PadChest                 & 101,752      & -             & 2,641	         & -          & -             & -               & -             & -               \\
BIMCV-COVID19            & 45,525       & -             & 1,202         & -              & -             & -               & -             & -               \\
CheXpert                 & 45,491       & 181,619       & 1,112	         & 4,589              & -             & -               & -             & -               \\
OpenI                    & 3,252        & 3,628         & 85            & 92             & -             & -               & -             & -               \\
MIMIC-CXR                & 148,374      & 181,166       & 3,799         & 4,650          & -             & -               & -             & -               \\ \midrule
Total                    & 344,394      & 366,413       & 8,839         & 9,331          & 2,692         & 2,967           & 1,063         & 1,428           \\ \bottomrule
\end{tabular}}
\end{table*}

\subsection{Evaluation} \label{sec:task_a_evaluation}
We applied two types of metrics to evaluate different systems: \textit{n}-gram-based and model-based metrics. For the former, we adopted BLEU-4~\cite{papineni2002bleu} and ROUGE-L~\cite{lin2004rouge}, whereas for the latter, we adopted BERTScore~\cite{zhang2019bertscore}, F1-CheXbert~\cite{smit2020combining}, and F1-RadGraph~\cite{delbrouck2022improving}. To standardize the evaluation process, we used the same script from ViLMedic to evaluate all systems. By doing so, we avoid different teams using different versions or hyperparameters for a given metric -- for example, some existing studies use differing versions of BERTScore, leading to inconsistent score reporting.

\subsection{Results} \label{sec:task_a_results}
The automatic results for the Findings and Impression sections are shown in Tables~\ref{table:task_a_leaderboard_findings} and~\ref{table:task_a_leaderboard_impression}, respectively (Note: \textit{iHealth-Chile-1} did not submit scores for Impression generation, and thus is not included in Table~\ref{table:task_a_leaderboard_impression}). We congratulate \textit{e-Health CSIRO}, \textit{MAIRA}, and \textit{AIRI} for their outstanding performance on both Findings and Impression generation. It is also worth highlighting that the other teams (\textit{Gla-AI4BioMed}, \textit{SICAR}, \textit{CID}, \textit{iHealth-Chile-3\&2}, and \textit{iHealth-Chile-1}) designed novel solutions as well, providing insights for future research in this field beyond the competition. We also ran an evaluation using GREEN for the top 2 best-scoring systems (\textit{e-Health CSIRO} and \textit{MAIRA}) and recorded scores of 36.9 and 35.2, respectively, aligning with the leaderboard rankings\footnote{We adopted GREEN instead of the naive GPT-4 pairwise comparison since \citet{ostmeier_green_2024} found GPT-4 to have low correlation with expert preference.}.

\begin{table*}[ht]
\caption{RRG24 Leaderboard for the Findings Section}
\label{table:task_a_leaderboard_findings}
\centering
\resizebox{0.95\linewidth}{!}{
\begin{tabular}{@{}llcccccc@{}}
\toprule
                       &                        &                                            & \multicolumn{5}{c}{Automatic Evaluation Metrics $\uparrow$}                                                                                                   \\ \cmidrule(l){4-8} 
\multirow{-2.3}{*}{Rank} & \multirow{-2.3}{*}{Team} & \multirow{-2.3}{*}{Overall Score $\uparrow$} & BLEU-4                        & ROUGE-L                       & BERTScore                     & F1-CheXbert                   & F1-RadGraph                   \\ \midrule
1                      & e-Health CSIRO         & \cellcolor[HTML]{C0C0C0}\bfseries35.56              & \cellcolor[HTML]{C0C0C0}\bfseries11.68 & \cellcolor[HTML]{EFEFEF}26.16 & \cellcolor[HTML]{EFEFEF}53.80 & \cellcolor[HTML]{EFEFEF}57.49 & \cellcolor[HTML]{C0C0C0}\bfseries28.67 \\
2                      & MAIRA                  & \cellcolor[HTML]{EFEFEF}35.08              & \cellcolor[HTML]{EFEFEF}11.24 & \cellcolor[HTML]{C0C0C0}\bfseries26.58 & \cellcolor[HTML]{C0C0C0}\bfseries54.22 & \cellcolor[HTML]{C0C0C0}\bfseries57.87 & \cellcolor[HTML]{EFEFEF}25.48 \\
3                      & AIRI                   & 33.55                                      & 9.97                          & 25.82                         & 52.42                         & 54.25                         & 25.29                         \\
4                      & Gla-AI4BioMed          & 31.01                                      & 7.65                          & 24.35                         & 52.69                         & 46.21                         & 24.13                         \\
5                      & SICAR                  & 30.93                                      & 6.62                          & 23.66                         & 50.74                         & 49.00                         & 24.62                         \\
6                      & CID                    & 30.71                                      & 7.46                          & 23.30                         & 50.89                         & 50.47                         & 21.45                         \\
7                      & iHealth-Chile-3\&2     & 23.38                                      & 4.81                          & 15.96                         & 44.03                         & 33.69                         & 18.41                                          \\
8                      & iHealth-Chile-1        & 20.83                                      & 6.46                          & 20.51                         & 49.23                         & 9.35                          & 18.59                         \\ \bottomrule
\end{tabular}}
\end{table*}

\begin{table*}[ht]
\caption{RRG24 Leaderboard for the Impression Section}
\label{table:task_a_leaderboard_impression}
\centering
\resizebox{0.95\linewidth}{!}{
\begin{tabular}{@{}llcccccc@{}}
\toprule
                       &                        &                                            & \multicolumn{5}{c}{Automatic Evaluation Metrics $\uparrow$}                                                                                                   \\ \cmidrule(l){4-8} 
\multirow{-2.3}{*}{Rank} & \multirow{-2.3}{*}{Team} & \multirow{-2.3}{*}{Overall Score $\uparrow$} & BLEU-4                        & ROUGE-L                       & BERTScore                     & F1-CheXbert                   & F1-RadGraph                   \\ \midrule
1                      & e-Health CSIRO         & \cellcolor[HTML]{C0C0C0}\bfseries35.28              & \cellcolor[HTML]{C0C0C0}\bfseries12.33 & \cellcolor[HTML]{EFEFEF}28.32 & \cellcolor[HTML]{EFEFEF}50.94 & \cellcolor[HTML]{C0C0C0}\bfseries56.97 & \cellcolor[HTML]{C0C0C0}\bfseries27.83 \\
2                      & MAIRA                  & \cellcolor[HTML]{EFEFEF}34.06              & \cellcolor[HTML]{EFEFEF}11.66 & \cellcolor[HTML]{C0C0C0}\bfseries28.48 & \cellcolor[HTML]{C0C0C0}\bfseries51.62 & \cellcolor[HTML]{EFEFEF}53.27 & \cellcolor[HTML]{EFEFEF}25.26 \\
3                      & AIRI                   & 32.98                                      & 10.91                         & 27.46                         & 49.55                         & 52.32                         & 24.67                         \\
4                      & SICAR                  & 30.73                                      & 8.03                          & 24.29                         & 47.15                         & 52.73                         & 21.46                         \\
5                      & Gla-AI4BioMed          & 30.46                                      & 9.60                          & 25.27                         & 48.60                         & 46.74                         & 22.10                         \\
6                      & CID                    & 25.21                                      & 7.13                          & 20.41                         & 43.67                         & 39.64                         & 15.19                         \\
7                      & iHealth-Chile-3\&2     & 17.30                                      & 1.66                          & 10.21                         & 37.21                         & 25.82                         & 11.58                         \\ \bottomrule
\end{tabular}}
\end{table*}

\subsection{Descriptions of Systems} \label{sec:task_a_systems}
\subsubsection{e-Health CSIRO}
\textbf{e-Health CSIRO}~\cite{e-Health-2024-rrg24} integrated entropy regularization into self-critical sequence training to help maintain a higher entropy in the token distribution, preventing overfitting to common phrases and ensuring a broader exploration of the vocabulary during training. They applied this to a multimodal language model with RadGraph as the reward. Additionally, their model incorporated several other features: (i) the use of type embeddings to differentiate between Findings and Impression section tokens; and (ii) the use of a non-causal attention mask for image embeddings and a causal mask for report token embeddings.

\subsubsection{MAIRA}
\textbf{MAIRA}~\cite{maira-2024-rrg24} combined a CXR-specific image encoder with a pre-trained LLM (Vicuna-7B-v1.5) via a multi-layer perceptron (MLP) adapter of 4 layers. The image encoder is a ViT-B model that leverages DINOv2, a state-of-the-art self-supervised learning method. Both the LLM and the adapter are fine-tuned in a single stage training setup for RRG. Their results indicated that joint training for Findings and Impression prediction improves the metrics for Findings generation. Additionally, incorporating lateral images alongside frontal images further enhances all metrics. They showed that scaling the model size from Vicuna-7B to Vicuna-13B also improves metrics. To handle multiple predictions for a study (as each study can have multiple frontal and/or lateral images), they utilized GPT-4 to select the best report.

\subsubsection{AIRI}
\textbf{AIRI}~\cite{airi-2024-rrg24} utilized the LLaVA framework, where the vision encoder is a DINOv2 trained on medical data and the language decoder is a specialized biomedical LLM. They used the same model to generate both Impressions and Findings with different prompts: ``Write findings for this X-ray.'' or ``Write impression for this X-ray.''. The system prompt from LLaVA-Med~\cite{li2024llava} was also used.

\subsubsection{Gla-AI4BioMed}
\textbf{Gla-AI4BioMed}~\cite{gla-2024-rrg24} leveraged the Vicuna-7B architecture and integrated a CLIP image encoder with a fine-tuned LLM. The model underwent a two-stage training process, whereby chest X-ray features are initially aligned with the language model, and said model is subsequently fine-tuned for report generation. The model processed multiple images simultaneously by stitching them together, mimicking the workflow of radiology professionals.

\subsubsection{SICAR}
\textbf{SICAR}~\cite{sicar-2024-rrg24} incorporated the SigLIP vision encoder and the Phi-2-2.7B language model to train an efficient RRG model. They also implemented a novel two-stage post-processing pipeline. They first enhanced the readability and clarity of the reports, then cross-verified the model outputs by integrating X-Raydar, an advanced X-ray classification model, addressing false negatives.

\subsubsection{CID}
\textbf{CID}~\cite{cid-2024-rrg24} proposed a novel paradigm for incorporating graph structural data into the RRG model. Their approach involved predicting graph labels based on visual features and subsequently initiating the decoding process through a template injection conditioned on the predicted labels. These results provided preliminary evidence for the feasibility of this new approach, which warrants further exploration in the future.

\subsection{iHealth-Chile-3\&2}
\textbf{iHealth-Chile-3\&2}~\cite{i-Health-32-2024-rrg24} focused on exploring various template-based strategies using predictions from multi-label image classifiers as input, which was inspired by prior work on template-based report generation. Two approaches were explored: (i) a straightforward implementation from~\citet{pino2021clinically} directly; and (ii) replacing the fully connected layer with an attention-based pooling mechanism conditioned on a fact embedding.

\subsection{iHealth-Chile-1}
\textbf{iHealth-Chile-1}~\cite{i-Health-1-2024-rrg24} developed a new strategy for in-context learning. Their system is formed using a vision-encoder, a vision-language connector or adapter, and a LLM able to process text and visual embeddings. They also designed an enriched prompt by combining a standard instruction (``Write the finding section of a chest x-ray radiology report'') with reports generated by a multi-label classifier and a group of template sentences.

\subsection{Limitations \& Challenges}\label{sec:task_a_limitations}
The evaluation for medical text generation is challenging due to its domain-specific characteristics, making it difficult to measure performance as it relates to clinical utility. This challenge leveraged common metrics that are used by existing RRG studies. Unfortunately, these evaluations may be limited when considering the real-world clinical impact of the submitted systems.

\section{``Discharge Me!'': Discharge Summary Generation} \label{sec:task_b}
``Discharge Me!'' was hosted on Codabench~\cite{Xu_2022}, an open source platform used to organize various tasks and benchmarks. A total of 211 submissions was received from across 16 teams.

\subsection{Data} \label{sec:task_b_data}
Participants were provided a dataset derived from the MIMIC-IV-Note module~\cite{johnson2023mimic}. The modified and filtered dataset included 109,168 hospital admissions from the Emergency Department (ED), split into four sets (Training, Validation, Phase I Test, and Phase II Test)~\cite{xu_discharge_nodate}. Each visit includes chief complaints and diagnosis codes (either ICD-9 or ICD-10) documented by the ED\footnote{We assume ED diagnosis codes are available to the discharging clinician as ED documentation is likely to be complete at the time of discharge in most cases. However, we acknowledge that ICD codes may not necessarily be  finalized, so they will be removed in future iterations of the shared task.}, at least one radiology report, and a discharge summary with both BHC and Discharge Instructions sections. 

The generation targets for the BHC were extracted from the full discharge notes using a complex regular expression strategy that searched for relevant section headers and new-line formatting characters. A similar strategy was used for Discharge Instructions; however, given that this section is usually located at the end of a discharge note as its very last section, extraction was more trivial. Samples where the extracted length of either section was shorter than 10 words were removed and deemed invalid. The complete breakdown of the dataset is available in Table~\ref{table:task_b_breakdown}.

\begin{table*}[ht]
\caption{Dataset Breakdown for ``Discharge Me!''}
\label{table:task_b_breakdown}
\centering
\resizebox{0.8\linewidth}{!}{
\begin{tabular}{@{}lccccc@{}}
\toprule
Item & Total Count & Training & Validation & Phase I Test & Phase II Test\\
\midrule
Hospital Visits & 109,168 & 68,785 & 14,719 & 14,702 & 10,962\\
Discharge Summaries & 109,168 & 68,785 & 14,719 & 14,702 & 10,962\\
Radiology Reports & 409,359 & 259,304 & 54,650 & 54,797 & 40,608\\
ED Stays \& Chief Complaints & 109,403 & 68,936 & 14,751 & 14,731 & 10,985\\
ED Diagnoses & 218,376 & 138,112 & 29,086 & 29,414 & 21,764\\
\bottomrule
\end{tabular}}
\end{table*}

Participants were allowed to incorporate external datasets, either publicly available or proprietary, as well as link additional patient data from other MIMIC-IV modules. Additionally, with the exception of the test dataset, participants were given the flexibility of using all or part of the provided dataset in any combination as they see fit.

\subsection{Evaluation} \label{sec:task_b_evaluation}
\subsubsection{Automatic Scoring} \label{sec:task_b_evaluation_automatic}
Automatic scoring took place on Codabench with a Python 3.9 environment. A hidden subset of 250 samples from the test datasets of the respective phases was used to evaluate the submissions. The metrics for this task were based on a combination of textual similarity (\textit{n}-gram-based lexical metrics) and factual correctness of the generated text. Specifically, we considered the following metrics to automatically score submissions: BLEU-4~\cite{papineni2002bleu}, ROUGE-1/-2/-L~\cite{lin2004rouge}, BERTScore~\cite{zhang2019bertscore}, METEOR~\cite{banerjee_meteor_2005}, AlignScore~\cite{zha-etal-2023-alignscore}, and MEDCON~\cite{vanveen2024clinical}.

Initially, submissions were scored on both target sections separately (BHC \& Discharge Instructions). The mean across all test samples were computed for each metric, resulting in several performance scores for each of the two target sections (not reported on the leaderboard). Then, for each metric, we took the mean of the scores for each of the two target sections (reported under the metric name on the leaderboard). Finally, we computed the mean once again over all the metrics to arrive at a final overall system score (reported as Overall Score on the leaderboard). 

For instance, given $N$ samples, suppose $s$ is defined as the score for a given sample for a given metric, then the mean across all samples for a particular target section, $S$, would be calculated by:
\begin{equation}
S=\sum_{1}^{N}(s_{i})/N
\end{equation}
We then calculated $\beta$, the mean of a given metric over both target sections, for each of the 8 metrics using:
\begin{equation}
\beta=(S_{BHC}+S_{DischargeInstructions})/2
\end{equation}
Finally, the overall system score was calculated by taking the mean of the 8 $\beta$ values:
\begin{equation}
Overall=\sum_{1}^{8}(\beta_{i})/8
\end{equation}

\subsubsection{Clinician Scoring} \label{sec:task_b_evaluation_clinician}
At the end of the competition, the submissions from the top 6 best-scoring teams were reviewed by a group of six clinicians with diverse experiences in a broad range of specialties (two adult hospitalists, two clinical informatics fellows trained in pediatrics, a neurosurgeon, and a radiologist). Generated sections were evaluated for their completeness, correctness, and readability, as well as in a holistic comparison against the reference target sections (ground truth). In particular, completeness evaluates whether the generated text captures the clinically important information available in the reference text. In cases where there is inaccurate information, correctness specifies whether and how likely this mistake would lead to unintended impacts in future care. Readability was only evaluated by the clinicians for the BHC section as the intended audience of the Discharge Instructions section is the patient. Finally, the holistic comparison aimed to capture overall clinician preference.

Clinicians were presented with the reference target sections and the generated target sections side-by-side on a web-based survey dashboard hosted via Streamlit. Additionally, the full discharge summary was available in case reviewers required further context. They were then presented with a series of multiple-choice questions capturing each of the above criteria in a scale from 1 to 5, where 1 was the most negative option, and 5 was the most positive option.

Each clinician was provided with generated samples from three teams for evaluation. To minimize recall bias, we presented the generated submissions from all three teams consecutively in a randomized order for one particular sample, before moving onto the next.

Each team's submission was evaluated by three separate clinician reviewers. Scores were averaged and several agreement and reliability scores were calculated, including Cohen's Kappa and Fleiss Kappa for interobserver agreement~\cite{mchugh_interrater_2012, landis_measurement_1977}, as well as the intraclass correlation coefficient (ICC)~\cite{liljequist_intraclass_2019}.

\subsection{Results} \label{sec:task_b_results}
\subsubsection{Automatic Evaluation} \label{sec:task_b_results_automatic}
Automatic scoring of the submissions took place on Codabench's platform using queues connected to independent compute workers hosted on GCP. The final leaderboard on the Phase II Test set is available in Table~\ref{table:task_b_leaderboard}.

A baseline performance was available for participants to benchmark their submissions. The baseline outputs were generated by a LLaMA-2-7B model fine-tuned on radiology reports from MIMIC-III~\cite{johnson_mimic-iii_2016}. While the system exhibited some clinical domain knowledge, it struggled due to the diverse formatting of discharge summaries, which greatly differed from that of the radiology reports in the training set. All submissions exceeded the baseline performance.

\begin{table*}[ht]
\caption{``Discharge Me!'' Automatic Scoring Leaderboard}
\label{table:task_b_leaderboard}
\resizebox{1\linewidth}{!}{
\begin{tabular}{ccccccccccc}
\toprule
\multirow{2.6}{*}{\makecell{Rank}} & \multirow{2.6}{*}{\makecell{Team}} & \multirow{2.6}{*}{\makecell{Overall Score $\uparrow$}} & \multicolumn{8}{c}{Automatic Evaluation Metrics $\uparrow$}\\ 
\cmidrule(lr){4-11}
& & & \makecell{BLEU-4} & \makecell{ROUGE-1} & \makecell{ROUGE-2} & \makecell{ROUGE-L} & \makecell{BERTScore} & \makecell{METEOR} & \makecell{AlignScore} & \makecell{MEDCON} \\
\midrule
1 & WisPerMed & \cellcolor[HTML]{C0C0C0}\bfseries0.332 & \cellcolor[HTML]{C0C0C0}\bfseries0.124 & \cellcolor[HTML]{C0C0C0}\bfseries0.453 & \cellcolor[HTML]{C0C0C0}\bfseries0.201 & \cellcolor[HTML]{C0C0C0}\bfseries0.308 & \cellcolor[HTML]{C0C0C0}\bfseries0.438 & \cellcolor[HTML]{C0C0C0}\bfseries0.403 & \cellcolor[HTML]{C0C0C0}\bfseries0.315 & \cellcolor[HTML]{C0C0C0}\bfseries0.411\\
2 & HarmonAI Lab at Yale & \cellcolor[HTML]{EFEFEF}0.300 & \cellcolor[HTML]{EFEFEF}0.106 & 0.423 & 0.180 & 0.284 & \cellcolor[HTML]{EFEFEF}0.412 & 0.381 & 0.265 & 0.353\\
3 & aehrc & 0.297 & 0.097 & 0.414 & \cellcolor[HTML]{EFEFEF}0.192 & \cellcolor[HTML]{EFEFEF}0.284 & 0.383 & \cellcolor[HTML]{EFEFEF}0.398 & 0.274 & 0.332\\
4 & EPFL-MAKE & 0.289 & 0.098 & \cellcolor[HTML]{EFEFEF}0.444 & 0.155 & 0.262 & 0.399 & 0.336 & 0.255 & \cellcolor[HTML]{EFEFEF}0.360\\
5 & UF-HOBI & 0.286 & 0.102 & 0.401 & 0.174 & 0.275 & 0.395 & 0.289 & \cellcolor[HTML]{EFEFEF}0.296 & 0.355\\
6 & de ehren & 0.284 & 0.097 & 0.404 & 0.166 & 0.265 & 0.389 & 0.376 & 0.231 & 0.339\\
7 & DCT\_PI & 0.277 & 0.092 & 0.401 & 0.158 & 0.256 & 0.378 & 0.363 & 0.247 & 0.320\\
8 & IgnitionInnovators & 0.253 & 0.068 & 0.370 & 0.131 & 0.245 & 0.360 & 0.314 & 0.215 & 0.324\\
9 & Shimo Lab & 0.248 & 0.063 & 0.394 & 0.131 & 0.252 & 0.351 & 0.312 & 0.210 & 0.276\\
10 & qub-cirdan & 0.221 & 0.024 & 0.377 & 0.106 & 0.205 & 0.300 & 0.332 & 0.174 & 0.254\\
11 & Roux-lette & 0.206 & 0.030 & 0.319 & 0.084 & 0.182 & 0.289 & 0.287 & 0.195 & 0.265\\
12 & UoG Siephers & 0.191 & 0.017 & 0.341 & 0.109 & 0.209 & 0.268 & 0.247 & 0.143 & 0.193\\
13 & mike-team & 0.188 & 0.022 & 0.290 & 0.076 & 0.163 & 0.258 & 0.294 & 0.182 & 0.223\\
14 & Ixa-UPV & 0.183 & 0.016 & 0.259 & 0.057 & 0.144 & 0.282 & 0.284 & 0.210 & 0.215\\
15 & MLBMIKABR & 0.170 & 0.039 & 0.210 & 0.092 & 0.131 & 0.186 & 0.306 & 0.205 & 0.191\\
16 & cyq & 0.104 & 0.002 & 0.197 & 0.016 & 0.106 & 0.179 & 0.106 & 0.132 & 0.091\\
\bottomrule
\end{tabular}}
\end{table*}

\subsubsection{Clinician Evaluation} \label{sec:task_b_results_clinician}
Overall clinician review scores are available in Table~\ref{table:task_b_clinician_overall}, and the specific rankings for the BHC and Discharge Instructions sections are shown in Tables~\ref{table:task_b_clinician_bhc} and~\ref{table:task_b_clinician_di}, respectively (mean clinician scores are provided, along with their constituent scores in brackets). Interestingly, the rankings for the overall clinician review exactly reflected that of the automatic evaluation using the reported metrics.

Figure~\ref{fig:heatmap} illustrates the interobserver agreement between pairwise clinicians based on the Cohen's Kappa statistic calculated for common submissions reviewed. As not all clinicians reviewed the same subset of submissions, a statistic could not be calculated for all reviewers (\textit{i.e.,} reviewer \#5 and \#6 did not have any submissions in common). There was rather poor agreement between most clinicians, likely due to subjective aspects of the evaluation and varying clinician preference during the holistic comparison.

However, the Fleiss Kappa value indicated that the reviews for the top 6 best-scoring submissions, where each submission was reviewed by 3 individual clinicians, exhibited substantial to almost perfect agreement (Table~\ref{table:task_b_clinician_overall}). Moderate reliability was also observed for the review methodology, as inferred from the presented range of ICC values.

\begin{table}[ht]
\caption{``Discharge Me!'' Clinician Scoring Leaderboard}
\label{table:task_b_clinician_overall}
\resizebox{1\linewidth}{!}{
\begin{tabular}{ccccc}
\toprule
\makecell{Rank} & \makecell{Team} & \makecell{Average $\uparrow$} & \makecell{Fleiss Kappa} & \makecell{Intraclass Corr.} \\
\midrule
1 & WisPerMed & \cellcolor[HTML]{C0C0C0}\bfseries3.375 & 0.781 & 0.336\\
2 & HarmonAI Lab at Yale & \cellcolor[HTML]{EFEFEF}2.903 & 0.944 & 0.656\\
3 & aehrc & 2.785 & 0.904 & 0.685\\
4 & EPFL-MAKE & 2.720 & 0.896 & 0.563\\
5 & UF-HOBI & 2.579 & 0.923 & 0.574\\
6 & de ehren & 2.335 & 0.908 & 0.740\\
\bottomrule
\end{tabular}}
\end{table}

\label{fig:heatmap}
\begin{figure}[htbp]
\centering
\resizebox{0.90\linewidth}{!}{
  \includegraphics{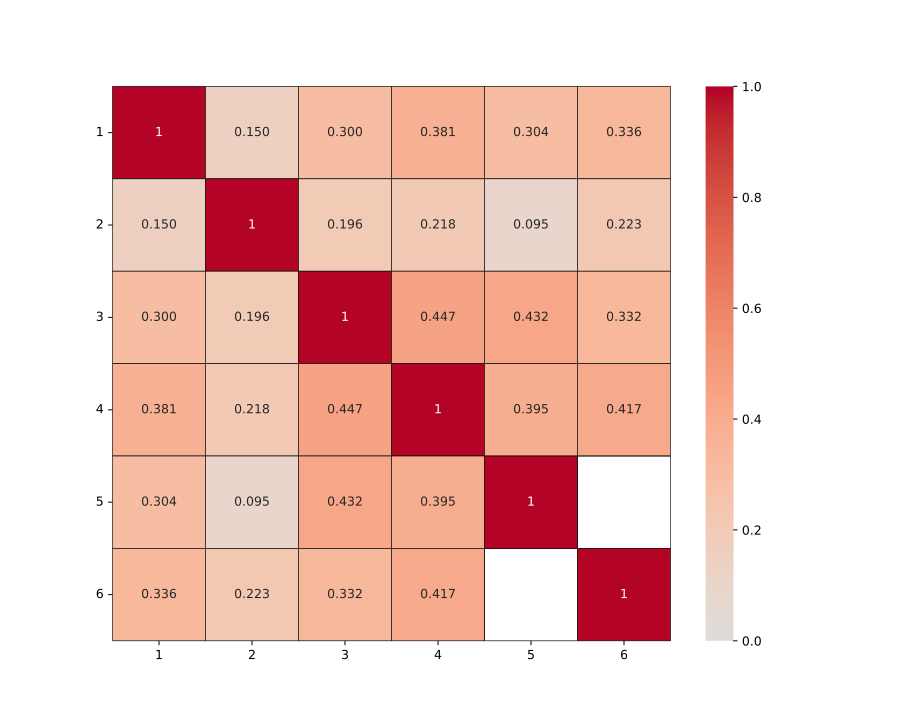}
  }
  \caption{Correlation heatmap visualizing interobserver agreement between clinician reviews. Cohen's Kappa scores were computed between pairwise clinicians based on the respective common submission(s) reviewed.}
\end{figure}

\subsubsection{Readability of Discharge Instructions} \label{sec:task_b_di_readability}
As the Discharge Instructions section is intended for patients who many not have medical training and knowledge of clinical acronyms, we decided to skip the clinician review and opted for an evaluation using common readability scores: the Flesch Reading Ease score and the Flesch–Kincaid Grade Level~\cite{friedman_systematic_2006}.
    
The writing of patient-targeting notes at an appropriate readability level is crucial as it directly relates to patient comprehension, engagement, and adherence to treatment plans post-discharge. Several healthcare institutes have placed recommendations on the readability of patient-facing material. Specifically, the National Institutes of Health (NIH) and American Medical Association (AMA) encourage a reading grade level of not higher than sixth-grade, while the Centers for Disease Control and Prevention (CDC) suggests a reading grade level of lower than eighth-grade~\cite{johnston_discharge_2018, cotugna_evaluation_2005, mccray_promoting_2005, burns_readability_2022}. 

A summary of the average readability metrics for the generated Discharge Instructions section is shown in Table~\ref{table:task_b_clinician_di}. The readability of most submissions hovered around a reading grade level of seventh-grade, with the exception of one team at around the ninth-grade. The reference sections had a Flesch Reading Ease score of 61.81 (± 11.92) and a Flesch–Kincaid Grade Level of 8.16 (± 2.12). As such, all evaluated systems were able to reasonably re-create the readability of the reference sections, with several able to generate Discharge Instructions that are more understandable and in-line with established guidelines.

\begin{table*}[ht]
\caption{``Discharge Me!'' Rankings based on Clinician Scoring of the Brief Hospital Course Section}
\label{table:task_b_clinician_bhc}
\resizebox{0.95\linewidth}{!}{
\begin{tabular}{ccccccc}
\toprule
\multirow{2.6}{*}{\makecell{Rank}} & \multirow{2.6}{*}{\makecell{Team}} & \multirow{2.6}{*}{\makecell{Average $\uparrow$}} & \multicolumn{4}{c}{Clinician Evaluation Criteria $\uparrow$}\\ 
\cmidrule(lr){4-7}
& & & \makecell{Completeness} & \makecell{Correctness} & \makecell{Readability} & \makecell{Holistic Comparison} \\
\midrule
1 & WisPerMed & \cellcolor[HTML]{C0C0C0}\bfseries3.29 & \cellcolor[HTML]{C0C0C0}\bfseries3.67 \tiny (4.08 3.16 3.76) & \cellcolor[HTML]{C0C0C0}\bfseries3.67 \tiny (4.20 3.40 3.40) & \cellcolor[HTML]{C0C0C0}\bfseries3.37 \tiny (3.76 3.40 2.96) & \cellcolor[HTML]{C0C0C0}\bfseries2.44 \tiny (2.96 2.60 1.76)\\
2 & EPFL-MAKE & \cellcolor[HTML]{EFEFEF}2.58 & 3.29 \tiny (3.28 3.20 3.40) & 2.83 \tiny (2.80 2.96 2.72) & 2.53 \tiny (2.88 2.56 2.16) & \cellcolor[HTML]{EFEFEF}1.65 \tiny (2.12 1.52 1.32)\\
3 & UF-HOBI & 2.49 & 2.48 \tiny (2.52 2.48 2.44) & \cellcolor[HTML]{EFEFEF}3.36 \tiny (3.48 3.28 3.32) & \cellcolor[HTML]{EFEFEF}2.71 \tiny (3.20 2.96 1.96) & 1.41 \tiny (1.96 1.20 1.08)\\
4 & HarmonAI Lab at Yale & 2.44 & \cellcolor[HTML]{EFEFEF}3.52 \tiny (3.32 3.64 3.60) & 2.59 \tiny (2.68 3.00 2.08) & 2.11 \tiny (2.36 2.00 1.96) & 1.53 \tiny (1.60 1.84 1.16)\\
5 & de ehren & 2.27 & 2.28 \tiny (2.36 2.32 2.16) & 2.99 \tiny (3.12 3.24 2.60) & 2.68 \tiny (2.72 2.84 2.48) & 1.12 \tiny (1.16 1.20 1.00)\\
6 & aehrc & 2.10 & 2.31 \tiny (2.24 2.52 2.16) & 3.05 \tiny (3.32 3.40 2.44) & 1.96 \tiny (2.16 1.80 1.92) & 1.09 \tiny (1.08 1.20 1.00)\\
\bottomrule
\end{tabular}}
\end{table*}

\begin{table*}[ht]
\caption{``Discharge Me!'' Rankings based on Clinician Scoring of the Discharge Instructions Section}
\label{table:task_b_clinician_di}
\resizebox{0.95\linewidth}{!}{
\begin{tabular}{cccccccc}
\toprule
\multirow{2.6}{*}{\makecell{Rank}} & \multirow{2.6}{*}{\makecell{Team}} & \multirow{2.6}{*}{\makecell{Average $\uparrow$}} & \multicolumn{3}{c}{Clinician Evaluation Criteria $\uparrow$} & \multicolumn{1}{c}{Flesch} & \multicolumn{1}{c}{Flesch-Kincaid}\\ 
\cmidrule(lr){4-6}
& & & \makecell{Completeness} & \makecell{Correctness} & \makecell{Holistic Comparison} & \makecell{Reading Ease} & \makecell{Grade Level}\\
\midrule
1 & aehrc & \cellcolor[HTML]{C0C0C0}\bfseries3.69 & 3.91 \tiny (3.80 4.40 3.52) & \cellcolor[HTML]{C0C0C0}\bfseries4.55 \tiny (4.52 4.48 4.64) & \cellcolor[HTML]{C0C0C0}\bfseries2.63 \tiny (2.48 3.24 2.16) & 62.05 \tiny (± 10.04) & 7.80 \tiny (± 1.76)\\
2 & HarmonAI Lab at Yale & \cellcolor[HTML]{EFEFEF}3.52 & \cellcolor[HTML]{C0C0C0}\bfseries4.27 \tiny (3.88 4.40 4.52) & 3.95 \tiny (3.84 3.88 4.12) & 2.36 \tiny (2.36 2.40 2.32) & 61.14 \tiny (± 14.52) & 8.60 \tiny (± 4.19)\\
3 & WisPerMed & 3.49 & \cellcolor[HTML]{EFEFEF}3.95 \tiny (4.36 3.36 4.12) & \cellcolor[HTML]{EFEFEF}4.00 \tiny (4.36 3.60 4.04) & \cellcolor[HTML]{EFEFEF}2.53 \tiny (2.48 2.76 2.36) & 63.35 \tiny (± 8.827) & 7.48 \tiny (± 1.53)\\
4 & EPFL-MAKE & 2.91 & 3.45 \tiny (3.28 3.36 3.72) & 3.41 \tiny (3.36 3.20 3.68) & 1.87 \tiny (2.20 1.64 1.76) & 58.72 \tiny (± 10.67) & 9.04 \tiny (± 1.81)\\
5 & UF-HOBI & 2.70 & 3.01 \tiny (2.60 3.24 3.20) & 3.29 \tiny (3.36 3.28 3.24) & 1.79 \tiny (2.00 1.84 1.52) & 66.73 \tiny (± 10.23) & 6.96 \tiny (± 1.57)\\
6 & de ehren & 2.43 & 2.81 \tiny (2.84 3.12 2.48) & 3.05 \tiny (3.36 3.12 2.68) & 1.41 \tiny (1.44 1.60 1.20) & 65.76 \tiny (± 8.706) & 7.28 \tiny (± 1.84)\\
\bottomrule
\end{tabular}}
\end{table*}

\subsection{Descriptions of Top Systems} \label{sec:task_b_systems}
A total of 12 system papers were received~\cite{wisper-2024-dm,yale-2024-dm,epfl-2024-dm,uf-2024-dm,shimo-2024-dm,ixa-2024-dm,qub-2024-dm,ehealth-2024-dm,uog-2024-dm,roux-2024-dm,gnitioninnovators-2024-rrg24,MLBMIKABR}. The top 6 best-scoring systems are detailed in this subsection.

\subsubsection{WisPerMed}
\textbf{WisPerMed}~\cite{wisper-2024-dm} investigated Dynamic Expert Selection (DES) consisting of a collection of LLMs fine-tuned and prompted for the task. They demonstrated that a DES system that chooses texts based on a specific length criteria performed the best on the given dataset. Thus, their objective with this strategy was to initially rank LLMs based on their archived overall scores. Subsequently, for each discharge summary, the generated sections (BHC \& Discharge Instructions) from the best model that had a word count within the range of 100 to 180 words was selected. If no model generated a block of text with a word count within this range, the text with the minimum word count greater than 70 words was selected. In cases where no piece of text met these criteria (\textit{i.e.,} shorter than 70 words), the text from the highest-ranked model was chosen. This approach emerged from the finding that longer pieces of medical text often led to hallucinations or repetitiveness.

\subsubsection{HarmonAI Lab at Yale}
The pipeline for \textbf{HarmonAI Lab at Yale}~\cite{yale-2024-dm} consisted of two BioBART-Large models. The one generating BHC sections was trained on all the preceding text prior to the BHC, while the Discharge Instructions model was trained on the BHC. The BHC model had an increased training dataset size due to shuffling and recombining the provided datasets. Default hyperparameter settings were largely used for training, with the exception of a lower learning rate. Models were trained for 2 epochs. For generation, a 4-beam search and limited repeats with an \textit{n}-gram size of 3 was employed. The minimum output length was set to 200 tokens based on the word count summary statistics and, and the maximum output token length was restricted to 1024 tokens due to the model specifications.

\subsubsection{aehrc}
\textbf{aehrc}~\cite{ehealth-2024-dm} used the content in the discharge summary note prior to the target sections as input context for both training and inference. To better handle the distinctions between the two sections, the team trained two separate models to generate the BHC and the Discharge Instructions. Their best model was based on PRIMERA, which is an encoder-decoder language model that is capable of handling extended input contexts and generating longer outputs. This model offered a slight edge over fine-tuning popular decoder-based LLMs at the 7/8B parameter-level with LoRA, and was also significantly faster at inference. Beam search with a size of 4 was used for decoding. 

\subsubsection{EPFL-MAKE}
\textbf{EPFL-MAKE}~\cite{epfl-2024-dm} mainly focused on the full-text available in the dataset as they believed that most of the useful information is hidden within. The text was used as an input into their system, which first extracted all sections that contained clinically useful information. The system then combined them into a new input. Some sections may have been removed if the new input was deemed too lengthy. The pre-processed input was then put into the medical LLM Meditron-7B, which is currently one of the top open-source medically pre-trained LLMs at the 7B level, to generate the BHC and Discharge Instructions sections. 

\subsubsection{UF-HOBI}
In their system, \textbf{UF-HOBI}~\cite{uf-2024-dm} employed two clinical LLMs that they have developed in their previous works, including an encoder-based model GatorTron~\cite{yang_large_2022} and a decoder-based model GatorTronGPT~\cite{peng_study_2023}. The team adopted GatorTron to extract clinical concepts from the discharge summary notes, and utilized GatorTronGPT to generate the BHC and Discharge Instructions sections. GatorTron, which was fine-tuned on the 2010 i2b2 Challenge Named Entity Recognition (NER) dataset, was used to extract three categories of concepts (“TEST”, “PROBLEM”, and “TREATMENT”) from the discharge summary and radiology reports for each visit. The extracted concepts were then used to form the generation model input. Two GatorTronGPT models were then trained using the P-tuning strategy for the generation of the two respective target sections. The model inputs were thus the concepts extracted from the various other sections.

\subsubsection{de ehren}
\textbf{de ehren} utilized Meerkat-7B-v1.0, a compact, instruction-tuned medical AI system renowned for its advanced medical reasoning capabilities. Meerkat excelled in various medical Question Answering (QA) benchmarks, notably achieving a score of 74.3 on MedQA. To further scrutinize its performance in long-form text generation and summarization tasks within the clinical domain, the team selectively extracted key sections from discharge summaries to fine-tune the model with regards to the model's attention window size.

\subsection{Limitations \& Challenges}\label{sec:task_b_limitations}
A primary concern was the risk of data leakage due to the release of the test sets with ground truth sections. To mitigate this, two test sets were released in two phases (one released at the start and one released much closer to the submission deadline), and the final evaluation was conducted on a hidden subset of 250 samples selected from the test datasets of the respective phases. This approach aimed to discourage participants from using the ground truth for model inference, or from optimizing systems for the tasks metrics throughout the entire duration of the competition. However, this method ultimately relies on the adherence of the participants to task guidelines.

The task also faced the challenge of dealing with inconsistently formatted free-text where ground truth generation targets are embedded within. The nature of clinical free-text can vary greatly, making it difficult to standardize inputs.

Furthermore, certain sections of the discharge summary appearing after the generation targets may not be reasonably available to the clinician at the time of discharge and the writing of the discharge summary. This presents a dilemma, as using such information would not accurately reflect the clinician's workflow. Although teams were reminded to justify any decisions made regarding the use of discharge summary sections, it was challenging to moderate this aspect.

Another limitation was the need to select discharge summaries of a reasonable length to make clinician review feasible. This selection process may introduce a bias, as longer or more complex summaries that could benefit from automated generation might be excluded. There was also plausible comparison bias during clinician review as clinicians were asked to review submissions that could have varied greatly in quality. However, we aimed to reduce this by randomizing the order in which submissions were presented to the clinicians.

\section{Conclusion}
As seen from the scores of the participating models for both tasks, there is great complexity in generating coherent, accurate, and clinically relevant free-text reports. Several factors contribute to this, including the inherent variability and nuance of natural language used in clinical settings.
 
It may be worthwhile to consider alternative approaches for fully automated report generation, such as by pre-processing reports into structured formats prior to AI generation. By breaking down the report generation process into more manageable tasks, generation systems may be able to achieve higher accuracy and coherence in their outputs~\cite{lederman_tasks_2022}. However, the standardization of formatting for these reports poses a significant challenge due to the diversity of writing styles and training among clinicians.

A previous study also explored the feasibility of generating hospital discharge summaries by tracing the source origin of medical expressions that make up the report~\cite{ando_is_2022}. Interestingly, the analysis found that a significant portion of the discharge summary originates from external sources rather than inpatient records, such as past clinical records, referral notes, and the expertise of the writing clinician. This suggests that an end-to-end generation pipeline would depend on advanced data retrieval and may ultimately require some form of manual clinician oversight.

Ultimately, we hope that this challenge will bolster the efforts of the biomedical natural language processing community in developing effective solutions for clinical text generation. We believe this task could form a solid foundation for future work on generating entire radiology reports or discharge summaries, which would help significantly reduce the time clinicians spend on administrative tasks and improve patient care quality.

\bibliography{custom}
\end{document}